\documentclass[twoside,leqno,twocolumn]{article}

\usepackage[letterpaper]{geometry}

\usepackage{ltexpprt}

\usepackage{float}
\usepackage[labelformat=simple]{subcaption}
\usepackage{adjustbox}
\usepackage{microtype}

\usepackage{algorithm}
\usepackage{algorithmic}
\usepackage{booktabs} 
\usepackage[hidelinks]{hyperref}
\usepackage{color}

\newcommand{\dsFont}[1]{{\fontfamily{qcr}\selectfont {#1}}}
\newcommand{\methodFont}[1]{{\fontfamily{lmtt}\selectfont {#1}}}

\usepackage{graphicx}
\usepackage{latexsym}
\usepackage{multirow}

\usepackage{amssymb}
\usepackage{mathtools}
\usepackage{diagbox}
\usepackage{bm}

\begin{document}

\newcommand\relatedversion{}

\title{\Large MISS: Multiclass Interpretable Scoring Systems}
\author{Michal K. Grzeszczyk\thanks{Sano Centre for Computational Medicine, m.grzeszczyk@sanoscience.org} \and Tomasz Trzciński \thanks{Warsaw University of Technology, IDEAS NCBR, Tooploox, tomasz.trzcinski@pw.edu.pl}
\and Arkadiusz Sitek \thanks{Massachusetts General Hospital and Harvard Medical School, asitek@mgh.harvard.edu}}

\date{}

\maketitle







\begin{abstract} \small\baselineskip=9pt In this work, we present a novel, machine-learning approach for constructing Multiclass Interpretable Scoring Systems (MISS) - a fully data-driven methodology for generating single, sparse, and user-friendly scoring systems for multiclass classification problems. Scoring systems are commonly utilized as decision support models in healthcare, criminal justice, and other domains where interpretability of predictions and ease of use are crucial. Prior methods for data-driven scoring, such as SLIM (Supersparse Linear Integer Model), were limited to binary classification tasks and extensions to multiclass domains were primarily accomplished via one-versus-all-type techniques. The scores produced by our method can be easily transformed into class probabilities via the softmax function. We demonstrate techniques for dimensionality reduction and heuristics that enhance the training efficiency and decrease the optimality gap, a measure that can certify the optimality of the model. Our approach has been extensively evaluated on datasets from various domains, and the results indicate that it is competitive with other machine learning models in terms of classification performance metrics and provides well-calibrated class probabilities. 

\end{abstract}

\section{Introduction}
\label{introduction}
Scoring systems are popular tools that are widely used in domains where conscious and fair decision-making is vital. Such models are characterized by the simplicity of adding small integers \cite{finlay2012credit}, the ease of use without electronic devices \cite{than2014development}, scarcity, interpretability, and transparency \cite{duwe2016sacrificing}. These attributes are the reason for their popularity in medicine \cite{gage2001validation}, criminal justice \cite{wang2022pursuit}, finance \cite{finlay2012credit}, and other domains.


Scoring systems \cite{ustun2016supersparse,zeng2017interpretable} require the user to conduct simple arithmetic operations (adding, subtracting and multiplying small integers) to compute the score conditionally on binary features (see Table \ref{tab:chads} and Appendix~\ref{appendix:scoring_systems} for examples). The score can be translated to class probability via a non-linear function (e.g. \textit{sigmoid}) or a pre-computed table. This enables one to choose the threshold for positive prediction depending on the case. 

Traditionally, those models are built in a semi-automated manner with the use of heuristics and expert knowledge (e.g. TIMI \cite{antman2000timi}), while others are created manually from scarce datasets (e.g. CHADS$_{2}$ \cite{gage2001validation}). Unfortunately, these approaches may violate many requirements and perform poorly in comparison to other models created on the same dataset \cite{ustun2019learning}. Also, there are often concerns regarding the fairness of such models \cite{kallus2018residual}. Optionally, these can be built using purely data-driven approaches \cite{ustun2016supersparse,ustun2019learning}. Ustun and Rudin developed Supersparse Linear Integer Model (SLIM), and its extension Risk-calibrated SLIM (RiskSLIM), methods utilizing Integer Programming (IP) to find optimal, small, integer coefficients for binary features to build scarce, interpretable models. RiskSLIM has been shown to perform well on many real-life datasets \cite{ustun2019learning,pajor2022effect,wang2022pursuit} while preserving simplicity and interpretability.
\begin{table}[t]
    \caption{CHADS$_{2}$ \cite{gage2001validation} is one of the many scoring systems utilized in healthcare. The sum of points evaluates to ischemic stroke probability as presented in the table.}
    \label{tab:chads}
    \centering
    \begin{tabular}{lc} \hline
    \textbf{Binary feature}               & \textbf{Points} \\ \hline
         Congestive heart failure history & 1 \\
         Hypertension                     & 1 \\
         Age $\geq$ 75                    & 1 \\
         Diabetes mellitus                & 1 \\
         Stroke or TIA symptoms           & 2 \\ \hline
    \end{tabular}
    \begin{tabular}{|@{}l|c|c|c|c|c|c|c|} \hline
        \textbf{\space Score}               & 0& 1& 2& 3& 4& 5& 6 \\ \hline
        \textbf{\space Pr. (\%)} & 1.9 & 2.8 & 4.0 & 5.9 & 8.5 & 12.5 & 18.2 \\ \hline
    \end{tabular}
    
\end{table}

RiskSLIM \cite{ustun2019learning} was proposed to solve binary classification problems like the detection of a single disease or the event of recidivism. However, real-life decision-making processes often require a multiclass analysis. There were approaches to adapt RiskSLIM or other scoring methods to multiclass settings although they were based on simplifying the multiclass problem. For instance, Rouzot {\em et al.}~\cite{rouzot2022learning} proposed a method to use RiskSLIM in a one-vs-rest (OvR) manner. While this solution can still be perceived as interpretable, it generates $K$ models for $K$ classes which makes it difficult to use in a feasible manner. Xu {\em et al.}~\cite{xu2020multi} have shown that the score produced by the scoring systems can be divided into ranges specified for each class. This method remains interpretable and feasible for multiclass problems that simplify the regression tasks - e.g. assessing the risk of developing the illness - the higher the score is, the more serious risk is predicted.

To address the limitations of previous methods, we propose Multiclass Interpretable Scoring Systems (MISS), a method to build a fully data-driven IP-based interpretable model (see example in Table~\ref{tab:example}). Our method creates a scoring system for the multiclass classification problem by optimizing the mixed-integer nonlinear program (MINLP). It is solved by minimizing the cross-entropy loss for maximization of Area Under the Receiver Operating Characteristics Curve (AUC) and calibration \cite{naeini2015binary} while penalizing the $l_{0}$-norm for the sparsity of the used features and restricting used coefficients to small integers. Each class in the given dataset has its own points assigned which allows to analyse the impact of every binary feature on the final prediction. Such formulation of the problem allows our method to be applicable to binary tasks as well, which broadens its utility. 
The score of each class for a specific sample is computed by adding points in rows corresponding to positive binary features of the MISS model (Table~\ref{tab:example}). The predicted class is the one with the highest score. The scores obtained by MISS can be used to compute class probabilities. To achieve this, the points for all classes have to be passed through the \textit{softmax} function. The MISS model developed here, is also paired with the \textit{optimality gap}, a measure certifying the optimality of the model or informing that the sufficiently accurate multiclass scoring system does not exist.

The main contributions of this paper are as follows:
\begin{enumerate}
  \item We present a data-driven method for learning multiclass scoring systems. Thanks to the utilization of IP, MISS can obey domain-specific constraints and is paired with the \textit{optimality gap} certifying the optimality of the model. Our method can also provide class probabilities.
  \item We develop methods for improving the performance of training MISS via novel dimensionality reduction with the Recursive Feature Aggregation (RFA) method and algorithmic improvements from the RiskSLIM \cite{ustun2019learning} adapted to the multiclass setting.
  \item We conduct experiments on several binary and multiclass, popular and real-life datasets \cite{detrano1989international,forina1991uci,street1993nuclear,fisher2011uci,Hurdman2012} and show that our method performs on par with several other machine learning methods. 
  \item We provide a publicly available code to train MISS with the CPLEX optimizer \cite{cplex2009v12} using a scikit-learn compatible API: \url{https://www.github.com/SanoScience/MISS}.
\end{enumerate}

\begin{table}[t]
    \caption{The example of trained MISS model for the \dsFont{iris} dataset. To classify a sample using this system, first, we determine which binary features are true, then sum corresponding points and \textit{bias}. The predicted class is the one with the highest number of points ({\em score}). Class probabilities can be calculated from scores using the \textit{softmax} function applied to the scores.}
     \label{tab:example}
    \resizebox{\columnwidth}{!}{
        \begin{tabular}{l*{3}{@{}c@{}}}\hline
                &  \multicolumn{3}{c}{Class} \\\cline{2-4}
                Binary feature & \textbf{setosa} \space& \textbf{versicolor} \space & \textbf{virginica}\\\hline
                $\bm{sepal\_length < 5.4}$& 1& -2& 1\\
                $\bm{4.8 \leq petal\_length}$& -5& 0& 5\\
                $\bm{0.8 \leq petal\_width < 1.75}$& -5& 5& 2\\
                $+$ \space $\bm{bias}$& 6& 3& 1 \\
                \hline
            Score:& = ....& = ....& = ....\\\hline
        \end{tabular}
    }
\end{table}

\section{Related work}
In this section, we describe existing methods for learning binary scoring systems. We show approaches to extend those methods to a multiclass setting and give examples of other multiclass interpretable models.

\textbf{Methods for building scoring systems.} Many methods were proposed for building scoring systems, e.g. via enforcing integer values on feature scores in support vector machines \cite{billiet2017interval}. The most popular approaches are based on a two-step process. First, the penalized logistic regression (LR) is trained on a dataset. Then, the real-valued coefficients are: rounded, rescaled and rounded, or assigned $\pm$1 depending on their sign. The last method is called \textit{Unit weighting} \cite{burgess1928factors} and was successfully applied in many cases \cite{antman2000timi,duwe2016sacrificing}. Liu \textit{et al.} \cite{liu2022fasterrisk} proposed a three-step framework called FasterRisk for learning binary risk scores by solving the sparse logistic regression problem, finding a pool of solutions with continuous coefficients and multiplying as well as rounding the coefficients to integer values.

MISS is related to other works that utilize IP to create sparse linear classifiers with integer coefficients \cite{ustun2016supersparse,ustun2019learning,verwer2019learning,haoran2021checklists}. This model can be viewed as the reformulation of RiskSLIM model proposed by Ustun amd Rudin \cite{ustun2019learning}. In our case, instead of learning a single coefficient set summed conditionally on binary features to compute the score for the probability assessment, we train simultaneously $K$ coefficient sets, where $K$ is the number of classes in the dataset. When the model is used, the scores do not have to be transformed through \textit{softmax} function, which is used during training, as the maximum score indicates the predicted class. The \textit{softmax} function is necessary only for assessing the probability. The problem is computationally harder as the number of coefficients and constraints of the IP problem increases linearly with the number of classes compared to \cite{ustun2019learning}.

\textbf{Multiclass scoring systems.} There are only few examples of methods that allow the generation of multiclass scoring systems. One of the most popular approaches to solving multiclass tasks is to reduce them to binary problems via one-vs-one or OvR. Such an approach was undertaken in \cite{rouzot2022learning}, but this method creates multiple scoring systems, which decreases its usability and interpretability. One scoring model for the multiclass setting was proposed in \cite{xu2020multi}. In this method, the classification scheme divides produced scores into multiple ranges - one per class. Such a scheme is only interpretable when there is a gradation between classes (e.g. \textit{none}, \textit{weak}, \textit{high}).

\textbf{Multiclass interpretable models.} There are many alternatives to LR and IP-based solutions that can be applied to multiclass classification problems in an interpretable way. Among the most popular algorithms, there are Decision Trees (DT) which are often applied to medical classification tasks \cite{sultana2018predicting}. RuleList \cite{proencca2020interpretable}, a multiclass extension of Rule Fit, is another example. 

\section{Method}
\label{sec:method}
In the following section, we define the problem of learning multiclass scores. We formulate the mixed integer nonlinear program for such systems and describe the process of finding the optimal solution by solving the surrogate problems. We also present the algorithmic improvements that enhance the performance of the optimizer. 

To create a MISS model, we use a dataset of \textit{n} training samples $(x_{i}, y_{i})_{i=0}^{n-1}$ where $x_{i} = [1, x_{i,1}, ..., x_{i,D}] \in \{0,1\}^{D+1}$ denotes a vector of \textit{D} binary features, $y_{i} \in \{0,1\}^{K}$ is a one-hot encoded label and \textit{K} is the number of classes. The per-class scores are a vector of \textit{K} values $s_{i} = [s_{i,0}, ..., s_{i,K-1}] = \left \langle x_{i},\lambda \right \rangle$ and are computed as a multiplication of $x_{i}$ and $\lambda$, where $\lambda \subseteq \mathbb{Z}^{(D+1)\times K}$ is a matrix of integer coefficients and $[\lambda_{0,0}, ..., \lambda_{0,K-1}]$ are per-class biases. In this setting, the $\lambda_{d,k}$ is the number of points added to the per-class score \textit{k} if the binary feature \textit{d} is positive. The per-class scores are transformed into a vector of class probabilities $r_{i}=[r_{i,0}, ..., r_{i, K-1}]$ with a \textit{softmax} function:
\begin{equation}
    r_{i,k}(s_{i}) = \frac{e^{s_{i,k}}}{\sum_{j=0}^{K-1} e^{s_{i,j}}}
\end{equation}

The benefit of such problem formulation is that, after model deployment, the \textit{softmax} function does not have to be applied. The predicted label $\hat{y_{i}}$ of sample $x_{i}$ is the class with the maximum score:
\begin{equation}
    \hat{y_{i}} = arg max_{0\leq k \leq K-1} s_{i,k}
\end{equation}

The multiclass scoring system  problem can be defined as a discrete optimization problem (mixed integer nonlinear program - MINLP) with the following form:
\label{eq:risk_score_problem}
\begin{align*}
     \underset{\lambda}{\min} \quad & l(\lambda)+C_{0}*B  \\
    s.t. \quad & \lambda \in \mathcal{L} \\
    & l(\lambda) = \frac{1}{n}\sum_{i=0}^{n-1}\sum_{k=0}^{K-1} -y_{i,k}*log(r_{i,k}(\langle x_{i}, \lambda \rangle)) \\
    & \mathcal{B}_{j} = (|| \lambda_{j,0} ||_{0} \vee ... \vee ||\lambda_{j,K-1} ||_{0})\\
    & B = \sum_{j=1}^{D}(\mathcal{B}_{j})\\
    & \mathcal{L} \subset \mathbb{Z}^{(D+1)} \\
    & C_{0} > 0 \\
\end{align*}

\begin{algorithm}[t]
   \caption{Recursive Feature Aggregation}
   \label{alg:RFA}
   \begin{algorithmic}
        \STATE {\bfseries Input:} \#  of features to choose $F$ \\
        \quad max coefficient value $\Lambda$ \\
        \quad max bias value $bias$\\
        \quad dataset $(x, y)$ \\
        \quad max features selected in every iteration $R^{max}$\\
        $features \gets \emptyset$ \\
        \IF{(\# features in $x$) $\leq F$}
         \STATE {\bfseries Output:}  features in $x$
        \ENDIF
        \FOR{$i \in F$}
         \STATE $features_{i} \gets$ MISS($x$, $y$, $\Lambda$, $bias$, $R^{max}$)
         \STATE Add ${features_{i}}$ to $features$
         \STATE Remove ${features_{i}}$ from $x$
        \ENDFOR
        \STATE {\bfseries Output:} $features$ \\
   \end{algorithmic}
\end{algorithm}

Here, $l(\lambda)$ is the normalized \textit{softmax cross-entropy} loss, $\mathcal{B}_{j}$ indicates whether at least one of \textit{j}-th feature coefficients is non-zero, $B$ is the $\ell_{0}$-seminorm penalizing the number of features used in the model (note that we do not penalize the bias terms), $\mathcal{L}$ is a set of user-allowed coefficient vectors and $C_{0}$ is a parameter for a trade-off between sparsity of the model and fit to the training dataset. MISS is a model that is an optimal solution to (\ref{eq:risk_score_problem}) and can be customized with the addition of user-specified constraints (see Appendix \ref{appendix:customization}).

\textbf{Optimization algorithm.}
The problem of solving this task is NP-hard, thus, MINLPs with integer-only constraints and $\ell_{0}$-regularization are computationally expensive. We find the solution to (\ref{eq:risk_score_problem}) with Lattice Cutting Plane Algorithm (LCPA) \cite{ustun2019learning} which is an improvement of other Cutting Plane Algorithms (CPA) \cite{kelley1960cutting}. CPA solves the optimization problems via the optimization of the surrogate mixed-integer program (MIP). The solution to MIP is removed from the region of search (the cut is added). With the addition of many cuts the optimization of the surrogate problem with multiple non-convex regions gets harder and CPA stalls. The LCPA solves MINLP with branch-and-bound (B\&B) search which splits feasible MINLP region into multiple. The surrogate linear program (LP) is solved over each region. The cuts are added only if an integer-feasible solution is found. This approach prevents the algorithm from stalling. We define the surrogate LP for the MISS in the following way:

\label{eq:surrogate_lp}
\begin{align*}
     \underset{L,\lambda,\beta}{\min} \quad & V  \\
    s.t. \quad & V = L + C_{0}R \\
    & 0 \leq K\beta_{j} - \sum_{k=0}^{K-1}\alpha_{j,k} \qquad j=1,...,D\\
    & 1 \geq K\beta_{j} - \sum_{k=0}^{K-1}\alpha_{j,k} \qquad j=1,...,D\\
    & R = \sum_{j=1}^{D}\beta_{j}\\
    & L \geq l(\lambda^{t}) + \langle \nabla l(\lambda^{t}), \lambda - \lambda^{t} \rangle \\
    & \qquad \qquad \qquad \qquad  t=1,...,c\\
    & \lambda_{j} \leq \Lambda_{j}^{max}\alpha_{j} \qquad j=1,...,D \\
    & \lambda_{j} \geq \Lambda_{j}^{min}\alpha_{j} \qquad j=1,...,D \\
    & \lambda \in \mathcal{R} \\
    \\
    & V \in [V^{min}, V^{max}] \\
    & L \in [L^{min}, L^{max}] \\
    & R \in [R^{min}, R^{max}] \\
    & \alpha_{j,k} \in [0, 1] \quad j=1,...,D; k=0,...,K-1\\
    & \beta_{j} \in [0, 1] \qquad j=1,...,D
\end{align*}

Here, $V$ is the objective value computed as the sum of $L$ loss and $\ell_{0}$ regularization term ($C_{0}R$), $l(\lambda^{t}) + \langle \nabla l(\lambda^{t}), \lambda - \lambda^{t} \rangle$ is one of a $c$ cuts at point $\lambda^{t}$ generated by LCPA from the value and gradient of the loss function, $\Lambda_{j}$ is user-provided feasible coefficient value, $\alpha_{j,k}$ is the $\ell_{0}$ binary indicator showing whether coefficient of $j$-th feature for $k$-th class is non-zero, $\beta_{j}$ is the $\ell_{0}$ binary indicator for  $j$-th feature indicating whether one of the coefficients $\alpha_{j,k}$ for all possible values of $k$ is non-zero and $R$ is the number of non-zero features bounded by the user-provided $R^{min}$ and $R^{max}$. The surrogate LP problem forces the optimizer to find the solution to the linear approximation of the loss function $l(\lambda)$. We follow the notation presented in \cite{ustun2019learning}.

\textbf{Algorithmic improvements.}
\label{seq:alg_improvements}
MISS has $K$-times more coefficients to find in comparison to binary RiskSLIM. While LCPA yields an optimal solution to MINLP in a finite time \cite{ustun2019learning} improvements can be made to accelerate the model's training process and decrease its \textit{optimality gap}. Here, we present methods for decreasing dimensionality at the beginning of the training and techniques to improve the lower bound (best-found mixed-integer solution) and the upper bound (best-found integer solution) during training.

The advantage of MISS is that it can find the best features to use in the scoring system during the optimization process. However, the more features in the dataset, the higher the computational time required to find the optimal model. To reduce the training time, we utilize the feature selection benefit of MISS and develop a Recursive Feature Aggregation method (Algorithm \ref{alg:RFA}). RFA, similarly to Recursive Feature Elimination \cite{guyon2002gene} trains models repeatedly on decreasing sets of features. In contrast to \cite{guyon2002gene} where feature importances are used to select the worst features to delete, we aggregate features selected by the MISS. RFA adds the best features to the selected features pool and removes them from the training set. In each iteration, we train the new MISS set $R^{max}$ to 1 to choose the best feature in the remaining set. RFA can be integrated within the MISS training process with the main model training parameters used as the input to feature selection models.

Bounds' improvement can be achieved with specialized heuristics and techniques. We extend \textit{Rounding Heuristic}, \textit{Polishing Heuristic}, and \textit{Bound Tightening} (\textit{BT}) procedures presented by Ustun and Rudin \cite{ustun2019learning} to multiclass application. \textit{Rounding Heuristic} rounds real-value solutions found by LCPA to find the locally optimal integer feasible solution. It rounds real values up and down contrary to naive rounding to the closest integers, which does not have to be optimal. \textit{Polishing Heuristic} polishes integer solutions via repeated increasing or decreasing of coefficients and ensures that the produced solution is \textit{1-opt} meaning that no better solution can be found by changing a single coefficient value. Both heuristics produce a stronger upper bound, reducing the time to find a good scoring system. On the other hand, \textit{Bound Tightening} increases the lower bound, which reduces the \textit{optimality gap}. \textit{BT} repeatedly updates constraints on $V^{min}$, $V^{max}$, $L^{min}$, $L^{max}$ and $R^{max}$ which removes non-optimal regions of B\&B.

\textbf{Optimality gap.} One of the benefits of an IP-based solution is that the model resulting from the training process is paired with the \textit{optimality gap}. During the B\&B search, the solver finds two types of solutions to the given problem: 1) a lower bound of the objective value (the minimal objective value for any feasible solution) i.e. $V^{min}$, 2) the best found integer-feasible solution $\lambda$ which gives the best found objective value $V^{max}=l(\lambda)$. Given $V^{min}$ and $V^{max}$ we can compute the \textit{optimality gap} of the produced MISS $\epsilon = 1-\frac{V^{min}}{V^{max}} \in [0, 1]$. When $V^{min}=V^{max}$ and $\epsilon=0$ this means that the solver has found the best possible solution to (\ref{eq:risk_score_problem}) and the model is \textit{certifiably optimal}.

Having the MISS paired with the \textit{optimality gap} gives the user more insight into the performance of the trained model and other possible solutions. The \textit{optimality gap} shows the worst-case difference between the produced model and the hypothetical optimal MISS that could be trained. Given the MISS, its objective value on the training set $L$ and the \textit{optimality gap} of $\epsilon$, we can predict that any other MISS has the objective value of at least $\left\lceil (1-\epsilon)L \right \rceil$ \cite{haoran2021checklists}. When $\epsilon$ is small, the objective value is close to the optimal objective value ($L$ of the model with $\epsilon=0$). The user can expect that no other MISS will perform better, when measured by performance on the training set, and can decide whether they are satisfied with the results. If not, the constraints given to the solver have to be relaxed. When $\epsilon$ is large, the user is informed that there might exist MISS with a lower objective value but the solver could not find it, requires more time or the dataset is too hard.

\textbf{Automated feature binarization.} Binarization is a process of transforming a dataset with $\hat{D}$ categorical and numerical features into a dataset with $D$ only-binary features. We one-hot encode the categorical features and treat missing values as an additional category. In our experiments, we follow  Feature Discretization RiskSLIM (FD-RiskSLIM) approach \cite{pajor2022effect} in which feature binarization is utilized within the hyperparameter optimization process. This method allows the application of many binarization techniques to find the one that fits best to the given problem and optimize its parameters. Many methods can be used by FD-RiskSLIM e.g. k-means or quantiles. 

\section{Experimental results}
\label{sec:experiments}

In what follows, we present the experimental setup for testing our method. We discuss results, different aspects of MISS (performance, interpretability, utility) and describe the limitations of our method.

\textbf{Datasets.} To develop and test MISS, we use three binary and five multiclass datasets (Table \ref{tab:datasets}). Some of them are popular e.g. \dsFont{iris} \cite{fisher2011uci}, others are real-life or clinical datasets e.g. \dsFont{heart} \cite{detrano1989international}. Only \dsFont{ph} and \dsFont{ph\_binary} datasets, containing information about Pulmonary Hypertension (PH) patients \cite{Hurdman2012} are not available online. We provide more information on data in Appendix \ref{appendix:datasets}. We binarize all categorical features. For the discretization of numerical features, we apply MDLP and k-bins discretizers from FD-RiskSLIM with strategy (k-means, uniform, quantiles) and a number of bins per feature subject to hyperparameter optimization. We report the best results for all binarization methods used. All numerical features are discretized in a non-overlapping way. We randomly oversample the minority class in the training set to counter class imbalances.

\begin{table*}[t!]
    \caption{We use three binary (B) and five multiclass (MC) datasets in Section \ref{sec:experiments}, containing $\hat{D}$ features before binarization. Here, $n$ is the number of samples in the dataset and $n_{ki}$ is the number of samples for class $i$.}
    \label{tab:datasets}
    \resizebox{\textwidth}{!}{
    \begin{tabular}{lcccccccll}\hline
        \textbf{Dataset}       & Type  & $\hat{D}$ & \textit{n}  & \textit{n$_{k0}$} & \textit{$n_{k1}$} & \textit{$n_{k2}$} & \textit{$n_{k3}$}  & Classification task            & Reference  \\ \hline
        \dsFont{diabetes} & B & 8 & 768 & 500 & 268 & - & -                 & Patient has diabetes & \cite{smith1988using}\\
        \dsFont{breast\_cancer} & B & 30 & 569 & 212 & 357 & - & -          & Patient has breast cancer & \cite{street1993nuclear}\\
        \dsFont{ph\_binary} & B & 35 & 353 & 66 & 287 & - & -               & Patient has PH & \cite{Hurdman2012}\\
        \dsFont{iris} & MC & 4 & 150 & 50 & 50 & 50 & -                  & Species of iris & \cite{fisher2011uci}\\
        \dsFont{wine} & MC & 13 & 178 & 59 & 71 & 48 & -                 & Cultivar of wine & \cite{forina1991uci}\\
        \dsFont{heart} & MC & 13 & 303 & 164 & 91 & 48 & -               & Patient has heart disease & \cite{detrano1989international}\\
        \dsFont{ph} & MC & 35 & 353 & 66 & 142 & 145 & -                 & Patient has arterial PH & \cite{Hurdman2012}\\    
        \dsFont{segmentation} & MC & 9 & 8068 & 1972 & 1858 & 1970& 2268 & Customer segmentation & \cite{segmentation_2021}\\
     \hline                                                  
    \end{tabular}
    }
\end{table*}

\textbf{Experimental setup.} We train \methodFont{MISS} with max points of 5 ($\lambda_{j,k} \in \{-5, ..., 5\}$ for every $j \in \{1, ..., D\}$ and $k \in \{0, ..., K-1\}$ ),  setting max bias to 20 ($[\lambda_{0,0}, ..., \lambda_{0,K-1}] \in \{-20, ..., 20\}^{K-1}$), maximum model size of $R^{max}=5$ ($B \leq 5$) and small sparsity penalization term $C_{0}=10^{-6}$ with a timeout of 90 minutes. We set $C_{0}$ to small values to favor smaller models over similarly accurate scoring systems which use more features. We apply algorithmic improvements presented in Section \ref{seq:alg_improvements} and set $F$, the number of features aggregated by RFA, to 15. We fit this method on a 3.00 GHz Intel Core i7 CPU and 32 GB RAM. 

For comparison, we use the following interpretable models: Unit weighting (\methodFont{UNIT}) \cite{burgess1928factors}, L1-penalized Logistic Regression (\methodFont{LR}), Decision Trees (\methodFont{DT}), \methodFont{RuleList} \cite{proencca2020interpretable}, One-vs-Rest FasterRisk (\methodFont{FasterRisk}) \cite{liu2022fasterrisk} and One-vs-Rest RiskSLIM (\methodFont{OvR-RiskSLIM}) \cite{ustun2019learning,rouzot2022learning}. We fit the ordinary RiskSLIM for binary classification datasets in the case of \methodFont{OvR-RiskSLIM} method. For multiclass datasets, we fit $K$ RiskSLIM models in a One-vs-Rest manner. We set parameters of RiskSLIM in a similar fashion as \methodFont{MISS} (max points of 5, max intercept of 20, max model size of 10, penalization term of $10^{-6}$ and a timeout of 20 minutes per each of $K$ models).  Additionally, we train RandomForest (\methodFont{RF}) and XGBoost (\methodFont{XGB}) \cite{chen2016xgboost} to create informal bounds on the performance metrics by non-interpretable but well-performing methods.

We optimize hyperparameters of all methods and binarization hyperparameters with Optuna, a hyperparameter optimization framework \cite{akiba2019optuna}. We run the optimization 10 times and maximize the weighted F1 metric during this process. We provide more details on search grids for all methods in Appendix \ref{appendix:hyperparameter}.

\textbf{Metrics.}  We conduct 5-Fold Stratified Cross-Validation (CV) and collect the F1, AUC, Expected Calibration Error (ECE) \cite{guo2017calibration} and \textit{optimality gap} as evaluation metrics. For \methodFont{OvR-RiskSLIM}, we compute the \textit{optimality gap} as the mean of gaps of all sub-models. We compute ECE with the \textit{uncertainty\_calibration} package \cite{kumar2019verified}. In a multiclass setting, we compute metrics in an OvR, sample-weighted manner \cite{zadrozny2002transforming}. 

\textbf{Ablation.} In Table \ref{tab:ablation}, we present the results of the ablation study over plain MISS and algorithmic improvements presented in the earlier sections. With the addition of \textit{Rounding} and \textit{Polishing Heuristics} together with \textit{BT} the performance of the model increases, however, the \textit{optimality gap} is very high. The dimensionality reduction with RFA brings a further improvement in the model performance. The \textit{optimality gap} drops steeply which means higher confidence in the produced models.

\begin{table}[t!]
    \caption{Ablation study on \dsFont{segmentation} dataset with MDLP discretization of the MISS performance with algorithmic improvements (\textit{Rounding Heuristic}, \textit{Polishing Heuristic}, \textit{Bound Tightening}) and RFA presented in Section \ref{sec:method}. The performance gain comes from the algorithmic improvements and dimensionality reduction with RFA which strongly reduces the optimality gap.}
    \label{tab:ablation}
    
    \centering
    \resizebox{\columnwidth}{!}{
    \begin{tabular}{ccc} \hline
                                    & AUC $\uparrow$      & \textit{optimality gap} $\downarrow$ \\ \hline
        \methodFont{MISS}    &  0.726  $\pm$ 0.005  &  0.40 $\pm$ 0.03     \\
         + heuristics \& BT&  0.731  $\pm$ 0.009  &  0.40 $\pm$ 0.04     \\
         + RFA                      &  0.741  $\pm$ 0.007 &  0.17 $\pm$ 0.01     \\ \hline
    \end{tabular}
    }
\end{table}

\textbf{Performance.} In Table \ref{tab:results_merged}, we present the results of 5-Fold CV on all tested methods and datasets. MISS achieves F1 and AUC which is comparable with other interpretable methods. Taking into account the sparsity and simplicity of generated models, which enhances their generalization possibilities, the overall performance is satisfactory. What is more, the produced class probabilities are well-calibrated often achieving the lowest ECE among all methods. MISS is also paired with the \textit{optimality gap}. For results on \dsFont{segementation} dataset with the AUC of 0.74 and \textit{optimality gap} of 0.17, we can expect that there might exist a MISS with the AUC of 0.78. The low \textit{optimality gap} indicates that the constraints (e.g. small integer coefficients) might be too strict and to achieve the better performance they should be relaxed. In the case of other interpretable models due to the lack of \textit{optimality gap} (apart from \dsFont{OvR-RiskSLIM}) we do not have the additional measure for deciding whether the problem is too hard or no better model exists.

\begin{table*}[t!]
    \caption{Performance metrics on three binary and five multiclass datasets. We report the mean and standard deviation on test sets of 5-Fold CV. The \textit{optimality gap} is given for the IP-based methods (we show the mean optimality gap of all sub-models for \methodFont{OvR-RiskSLIM}). We report results of \methodFont{RF} and \methodFont{XGB} as performance baselines.}
    \label{tab:results_merged}
    \resizebox{\textwidth}{!}{
    \begin{tabular}{ccccccccccc} 
        & & \multicolumn{2}{c}{BASELINES} & \multicolumn{6}{c}{INTERPRETABLE METHODS} \\ \cmidrule(r){3-4} \cmidrule(l){5-11}
        \textbf{Dataset} & \textbf{Metric} & \methodFont{RF} & \methodFont{XGB} & \methodFont{UNIT} & \methodFont{LR} & \methodFont{DT} & \methodFont{RuleList} & \methodFont{FasterRisk} & \methodFont{OvR-RiskSLIM} & \methodFont{\textbf{MISS}} \\ \hline
        & F1             &  0.67 $\pm$0.05 &  0.67 $\pm$0.05 &  0.56 $\pm$0.06 & \textbf{ 0.67 $\pm$0.04} &  0.64 $\pm$0.05 &  0.63 $\pm$0.07 &  \textbf{0.67 $\pm$0.05} &  0.63 $\pm$0.07 &  0.63 $\pm$0.05 \\
        \dsFont{diabetes} & AUC            &  0.81 $\pm$0.05 &   0.8 $\pm$0.05 &  0.75 $\pm$0.03 &  0.82 $\pm$0.05 &  0.77 $\pm$0.05 &  0.23 $\pm$0.06 &  0.81 $\pm$0.04 &  0.81 $\pm$0.04 &   0.8 $\pm$0.04 \\
        & ECE            &  0.12 $\pm$0.04 &  0.13 $\pm$0.03 &  0.18 $\pm$0.06 &  0.12 $\pm$0.02 &  0.17 $\pm$0.03 &  0.51 $\pm$0.05 &  0.11 $\pm$0.02 &   0.1 $\pm$0.02 &  0.12 $\pm$0.05 \\
        & opt\_gap &    - &    - &    - &    - &    - &    - &    - &  0.05 $\pm$0.01 &   0.1 $\pm$0.04 \\

                                                                      \hline
        & F1             &  0.98 $\pm$0.02 &  0.98 $\pm$0.01 &  0.96 $\pm$0.03 &  \textbf{0.97 $\pm$0.02} &  0.95 $\pm$0.02 &  0.96 $\pm$0.03 &  0.95 $\pm$0.02 &  0.95 $\pm$0.02 &  0.95 $\pm$0.01 \\
        \dsFont{breast\_cancer} & AUC            &  0.99 $\pm$0.01 &  0.99 $\pm$0.01 &  0.99 $\pm$0.01 &  0.99 $\pm$0.01 &  0.96 $\pm$0.01 &  0.97 $\pm$0.02 &  0.98 $\pm$0.01 &  0.98 $\pm$0.01 &  0.98 $\pm$0.02 \\
        & ECE            &  0.06 $\pm$0.01 &  0.05 $\pm$0.01 &  0.04 $\pm$0.01 &  0.04 $\pm$0.01 &  0.06 $\pm$0.01 &  0.03 $\pm$0.03 &  0.05 $\pm$0.02 &  0.05 $\pm$0.01 &  0.04 $\pm$0.02 \\
        & opt\_gap &    - &    - &    - &    - &    - &    - &    - &    1.0 $\pm$0.0 &  0.17 $\pm$0.04 \\

                                                                      \hline
        & F1             &  0.88 $\pm$0.15 &   0.9 $\pm$0.11 &  0.87 $\pm$0.13 &  0.88 $\pm$0.14 &   \textbf{0.9 $\pm$0.09} &   \textbf{ 0.9 $\pm$0.1} &  0.85 $\pm$0.18 &  0.89 $\pm$0.01 &  0.86 $\pm$0.16 \\
        \dsFont{ph\_binary} & AUC            &  0.91 $\pm$0.13 &  0.89 $\pm$0.14 &   0.89 $\pm$0.1 &  0.91 $\pm$0.11 &  0.78 $\pm$0.13 &  0.27 $\pm$0.36 &  0.86 $\pm$0.15 &   0.4 $\pm$0.15 &  0.88 $\pm$0.13 \\
        & ECE            &  0.18 $\pm$0.11 &  0.15 $\pm$0.12 &  0.16 $\pm$0.15 &  0.15 $\pm$0.15 &  0.15 $\pm$0.13 &  0.67 $\pm$0.34 &  0.17 $\pm$0.18 &  0.18 $\pm$0.03 &  0.15 $\pm$0.16 \\
        & opt\_gap &    - &    - &    - &    - &    - &    - &    - &  0.97 $\pm$0.06 &  0.19 $\pm$0.04 \\                                       
                                                                      \hline
        & F1        &  0.97 $\pm$0.03 &  0.97 $\pm$0.04 &  0.95 $\pm$0.05 &  \textbf{0.97 $\pm$0.03} &  \textbf{0.97 $\pm$0.06} &  \textbf{0.97 $\pm$0.04} &  0.96 $\pm$0.04 &  0.93 $\pm$0.07 &  \textbf{0.97 $\pm$0.03} \\
        \dsFont{iris} & AUC       &  0.99 $\pm$0.01 &  0.99 $\pm$0.01 &  0.99 $\pm$0.01 &  0.98 $\pm$0.02 &  0.98 $\pm$0.02 &  0.98 $\pm$0.02 &  0.99 $\pm$0.02 &  0.98 $\pm$0.02 &  0.99 $\pm$0.01 \\
        & ECE                &  0.04 $\pm$0.01 &  0.04 $\pm$0.02 &  0.17 $\pm$0.01 &  0.24 $\pm$0.03 &  0.02 $\pm$0.02 &  0.03 $\pm$0.01 &  0.02 $\pm$0.01 &  0.05 $\pm$0.04 &  0.02 $\pm$0.01 \\
        & opt\_gap     &    - &    - &    - &    - &    - &    - &    - &  0.02 $\pm$0.01 &  0.14 $\pm$0.02 \\                                                  
                                                                      \hline
        & F1        &  0.96 $\pm$0.03 &  0.97 $\pm$0.03 &  0.92 $\pm$0.06 &  0.96 $\pm$0.03 &  0.93 $\pm$0.03 &  0.93 $\pm$0.04 &  \textbf{0.97 $\pm$0.04} &  0.95 $\pm$0.03 &  0.93 $\pm$0.02 \\
        \dsFont{wine} & AUC       &  0.99 $\pm$0.01 &  0.99 $\pm$0.01 &  0.98 $\pm$0.02 &    1.0 $\pm$0.0 &  0.96 $\pm$0.03 &  0.95 $\pm$0.03 &  0.99 $\pm$0.01 &   1.0 $\pm$0.01 &  0.97 $\pm$0.02 \\
        & ECE                &  0.12 $\pm$0.03 &  0.04 $\pm$0.02 &  0.15 $\pm$0.03 &  0.04 $\pm$0.02 &  0.04 $\pm$0.01 &  0.05 $\pm$0.02 &  0.03 $\pm$0.01 &  0.04 $\pm$0.01 &  0.05 $\pm$0.03 \\
        & opt\_gap     &    - &    - &    - &    - &    - &    - &    - &   0.4 $\pm$0.26 &  0.86 $\pm$0.11 \\                     
                                                                      \hline

        & F1        &  0.68 $\pm$0.05 &  0.69 $\pm$0.02 &  0.64 $\pm$0.06 &  \textbf{0.69 $\pm$0.05} &  0.63 $\pm$0.07 &   0.6 $\pm$0.08 &  0.68 $\pm$0.07 &  0.68 $\pm$0.06 &  \textbf{0.69 $\pm$0.08} \\
        \dsFont{heart} & AUC       &  0.84 $\pm$0.03 &  0.83 $\pm$0.03 &  0.81 $\pm$0.05 &  0.85 $\pm$0.02 &  0.72 $\pm$0.06 &  0.79 $\pm$0.05 &  0.83 $\pm$0.03 &  0.82 $\pm$0.03 &  0.83 $\pm$0.03 \\
        & ECE                &   0.13 $\pm$0.0 &  0.15 $\pm$0.02 &  0.18 $\pm$0.03 &  0.14 $\pm$0.02 &  0.24 $\pm$0.04 &  0.14 $\pm$0.04 &  0.14 $\pm$0.02 &  0.16 $\pm$0.02 &  0.14 $\pm$0.03 \\
        & opt\_gap     &    - &    - &    - &    - &    - &    - &    - &  0.26 $\pm$0.05 &  0.32 $\pm$0.08 \\                                         
                                                                      \hline
         & F1        &  0.56 $\pm$0.11 &  0.56 $\pm$0.09 &   0.52 $\pm$0.1 &  \textbf{0.59 $\pm$0.06} &  0.53 $\pm$0.07 &  0.46 $\pm$0.12 &  0.54 $\pm$0.08 &  0.53 $\pm$0.08 &  0.46 $\pm$0.06 \\
        \dsFont{ph} & AUC       &  0.74 $\pm$0.07 &  0.75 $\pm$0.07 &  0.71 $\pm$0.06 &  0.74 $\pm$0.07 &  0.65 $\pm$0.02 &  0.47 $\pm$0.09 &  0.73 $\pm$0.07 &  0.73 $\pm$0.06 &  0.69 $\pm$0.04 \\
        & ECE                &  0.15 $\pm$0.05 &  0.17 $\pm$0.06 &  0.26 $\pm$0.07 &  0.19 $\pm$0.07 &  0.25 $\pm$0.04 &   0.3 $\pm$0.11 &  0.16 $\pm$0.06 &  0.18 $\pm$0.06 &  0.14 $\pm$0.08 \\
        & opt\_gap     &    - &    - &    - &    - &    - &    - &    - &  0.54 $\pm$0.02 &  0.22 $\pm$0.07 \\

                                                                    \hline
        & F1        &  0.52 $\pm$0.01 &  0.53 $\pm$0.01 &  0.42 $\pm$0.01 &  \textbf{0.51 $\pm$0.01} &  0.49 $\pm$0.01 &  \textbf{0.51 $\pm$0.01} &  0.49 $\pm$0.01 &  0.46 $\pm$0.01 &  0.44 $\pm$0.03 \\
        \dsFont{segmentation} & AUC       &  0.79 $\pm$0.01 &  0.79 $\pm$0.01 &  0.71 $\pm$0.01 &   0.77 $\pm$0.0 &  0.73 $\pm$0.01 &  0.46 $\pm$0.02 &  0.75 $\pm$0.01 &  0.74 $\pm$0.01 &  0.74 $\pm$0.01 \\
        & ECE                &  0.03 $\pm$0.01 &   0.03 $\pm$0.0 &  0.15 $\pm$0.01 &   0.03 $\pm$0.0 &   0.07 $\pm$0.0 &  0.19 $\pm$0.01 &   0.04 $\pm$0.0 &   0.04 $\pm$0.0 &   0.05 $\pm$0.0 \\
        & opt\_gap     &    - &    - &    - &    - &    - &    - &    - &   0.12 $\pm$0.0 &  0.17 $\pm$0.01 \\
                                                                
     
    \end{tabular}
    }
\end{table*}

\begin{figure}[t!]
\minipage{0.5\columnwidth}
  \includegraphics[width=\textwidth]{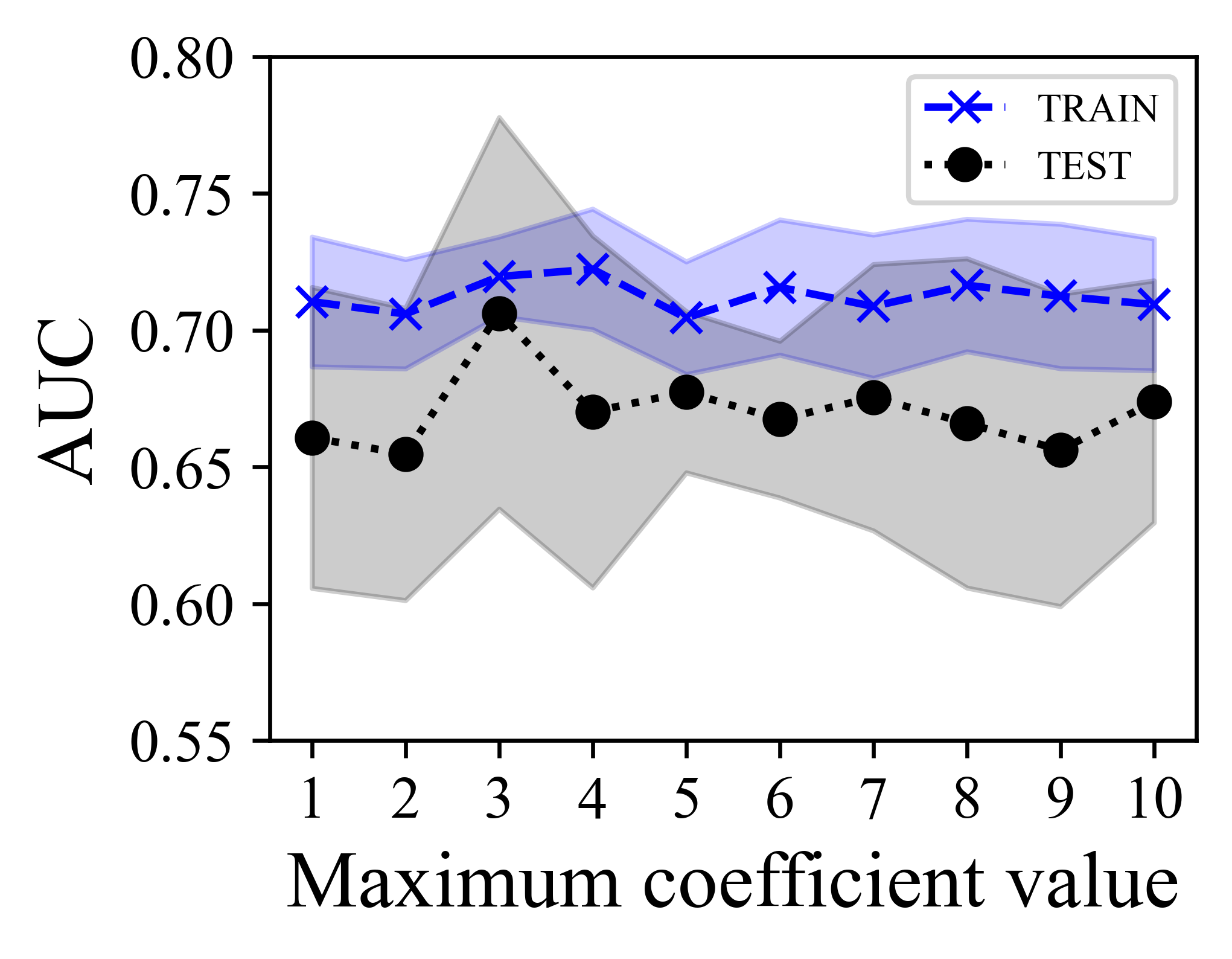}
\endminipage\hfill
\minipage{0.5\columnwidth}
  \includegraphics[width=\textwidth]{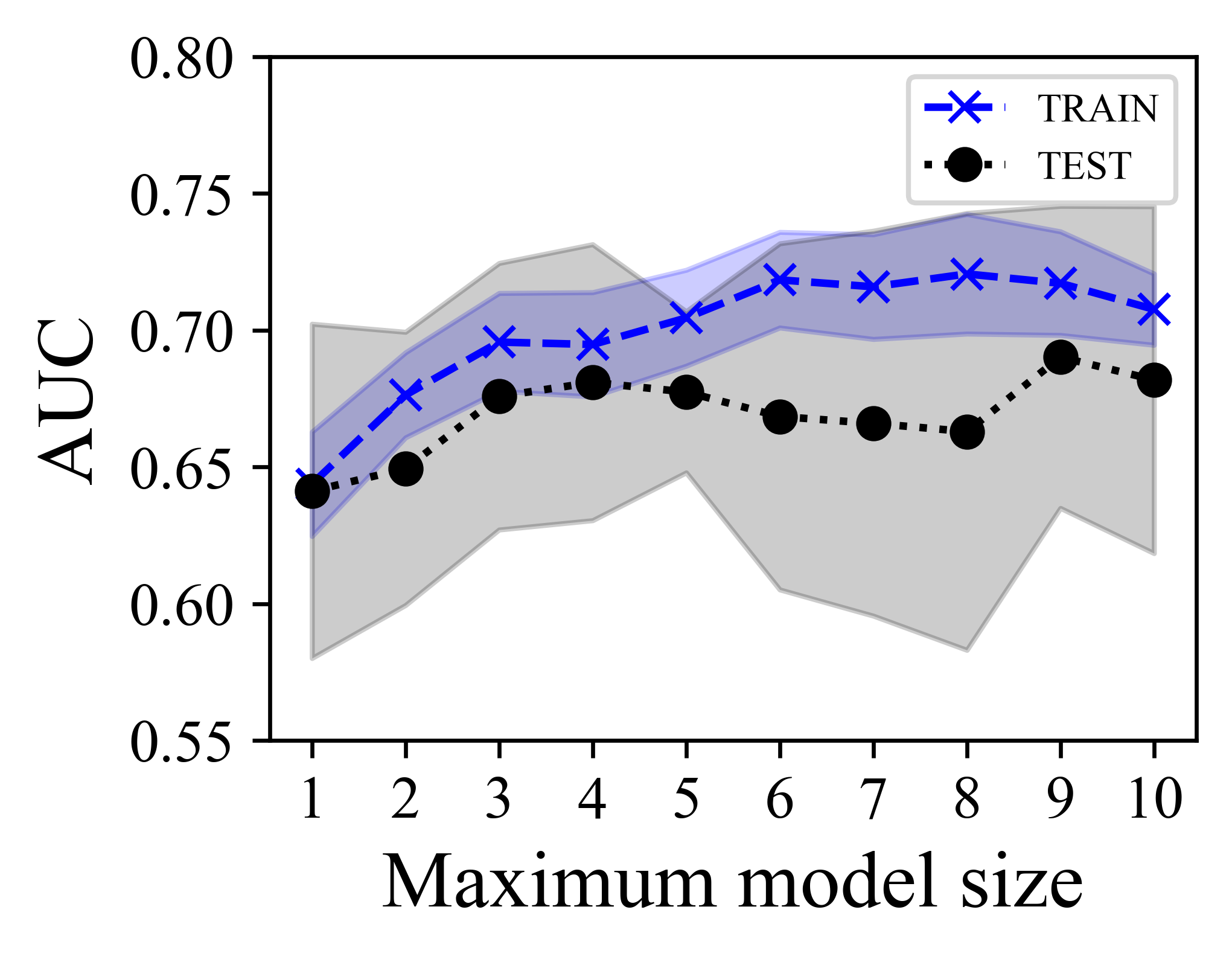}
\endminipage\hfill
\caption{Performance of MISS with $\Lambda^{max} \leq 10$ (left) and $R^{max} \leq 10$ (right) on \dsFont{ph} dataset.} 
\label{fig:charts}
\end{figure}

\textbf{Model size and max coefficient constraints.} In Figure \ref{fig:charts}, we present how the performance of MISS changes depending on the maximum value of the coefficient ($\Lambda^{max}$) and the maximum model size ($R^{max}$). The performance does not vary significantly with the changes of $\Lambda^{max}$, thus, we set $\Lambda^{max}$ to 5 to improve interpretability with low coefficient values. We set $R^{max}$ to 5, as above this value the model is less generalizable and is harder to use and understand.

\textbf{Interpretability.} The advantage of interpretable models is that the user can understand how the specific class was chosen, by following the prediction steps of the algorithm (splits in DT, multiplication of weights in LR etc.). The problem appears when the number of steps to produce the prediction is growing. LR is generally perceived as an interpretable algorithm, however, with a growing number of features the number of weights is increasing, thus making the prediction more complex. The number of non-zero weights can be diminished with regularization techniques, but the problem of understanding real-valued numbers with many decimal places persists. MISS has feature selection capabilities thanks to the penalization term used in the objective value and model-size constraints forcing the sparsity of the model. The restriction of coefficients to small integer values improves the interpretability of points assigned to each binary feature. Thanks to the binarization of features, the prediction steps are restricted to the summation of several numbers for every class and the comparison of the scores. All of those factors ensure that the model satisfying all given constraints is sparse, easy to understand, has good performance and calibration.

In our current implementation, the bias values for different classes exhibit variation, which may not be immediately transparent in terms of interpretability. However, the explanation is analogous to that of a traditional scoring model such as RiskSLIM. Specifically, when all binary features are false, the bias represents the class logits before applying the \textit{softmax} function. It is important to note that when all binary features are negative, this carries information about the system and is reflected in the bias. As a future avenue of research, we plan to explore an alternative approach where the bias for each class is constrained to be identical. While this approach may result in decreased performance due to stricter constraints, it offers another type of MISS.


In Table \ref{tab:ovr_vs_mc}, we present the comparison of three RiskSLIM models generated by OvR-RiskSLIM for \methodFont{heart} dataset with one model produced by MISS. Instead of using different features for every class, our method uses a single set of features for all classes. What is more, the per-class scores are comparable, and to conduct the classification, the sum of all points is the only requirement. In OvR-RiskSLIM, different features and intercept values restrict users from comparing per-class scores which have to be passed through \textit{sigmoid} function to normalize values and compute probabilities.

It is also important to mention that users of the MISS method may be tempted to interpret features with positive coefficients as a trigger for specific classes (e.g. diseases or recidivism). However, we advise not to treat the chosen features as the causal factors of the predicted class. For example, in binary classification, the positive coefficient of one class might indicate that the positive binary feature is rarely found in the second. This does not explain the roots of the first class. The values of coefficients are dependent on the optimization algorithm and dataset, and may vary between executions on different CPUs or timeouts given for every model.

\begin{table}[t!]
\caption{Comparison of scoring systems generated by OvR-RiskSLIM (\ref{tab:example_ovr0}, \ref{tab:example_ovr1}, \ref{tab:example_ovr2}) with MISS created by our method (\ref{tab:example_mcriskslim}) on the \dsFont{heart} dataset. The models discriminate between \textit{No Heart Disease} (HD), \textit{Mediocre HD} and \textit{Severe HD} classes. They achieve AUC of 0.85 and 0.87 respectively. Note, that OvR-RiskSLIM utilizes different features per sub-model and between-class coefficients are not directly comparable.}
\label{tab:ovr_vs_mc}
\centering
    \begin{subtable}[t]{0.49\columnwidth}
        \caption{\textit{No HD} RiskSLIM}
        \adjustbox{width =\textwidth}{
        \begin{tabular}{lc}\hline
        \textbf{Binary feature} & \textbf{Points}\\\hline
            $\bm{cp=asymptomatic}$ & -2\\
            $\bm{restecg=normal}$ & 1\\
            $\bm{slope=upsloping}$ & 1\\
            $\bm{thal=reversable defect}$ & -2\\
            $\bm{ca=0.0}$ & 2\\
            \hline
            Score: & = ....\\\hline
            \multicolumn{2}{c}{\textbf{Probability \textit{No HD}:} $\frac{1}{1 +  exp(-1 - score)}$}\\\hline
            \end{tabular}
        }
        \label{tab:example_ovr0}
    \end{subtable}
    \begin{subtable}[t]{0.49\columnwidth}
        \caption{\textit{Mediocre HD} RiskSLIM}
        \adjustbox{width =\textwidth}{
        \begin{tabular}{lc}\hline
            \textbf{Binary feature} & \textbf{Points}\\\hline
            $\bm{exang=False}$ & -1\\
            $\bm{thal=normal}$ & -1\\
            $\bm{ca=2.0}$ & -1\\
            $\bm{56.0 \leq age \leq 77.0}$ & 1\\
            $\bm{0.0 \leq oldpeak < 2.0}$ & 1\\
            \hline
            Score: & = ....\\\hline
            \multicolumn{2}{c}{\textbf{Prob. \textit{Mediocre HD}:} $\frac{1}{1 +  exp(-1 - score)}$}\\\hline
        \end{tabular}
        }
        \label{tab:example_ovr1}
    \end{subtable}
    
    \begin{subtable}[t]{0.49\columnwidth}
        \caption{\textit{Severe HD} RiskSLIM}
        \adjustbox{width =\textwidth}{
        \begin{tabular}{lc}\hline
            \textbf{Binary feature} & \textbf{Points}\\\hline
            $\bm{cp=non-anginal}$ & -2\\
            $\bm{slope=upsloping}$ & -2\\
            $\bm{thal=reversable defect}$ & 1\\
            $\bm{ca=0.0}$ & -2\\
            $\bm{0.0 \leq oldpeak < 2.0}$ & -1\\
            \hline
            Score: & = ....\\\hline
            \multicolumn{2}{c}{\textbf{Probability \textit{Severe HD}:} $\frac{1}{1 +  exp(1 - score)}$}\\\hline
        \end{tabular}
        \label{tab:example_ovr2}
        }
    \end{subtable}

    \begin{subtable}[t]{\columnwidth}
        \caption{MISS for all \textit{HD} classes}
        \centering
        \adjustbox{width =\textwidth}{
        \begin{tabular}{l*{3}{c}}\hline
        &  \multicolumn{3}{c}{Class} \\\cline{2-4}
                Binary feature &  \textbf{No HD}& \textbf{Mediocre HD}& \textbf{Severe HD}\\\hline
            $\bm{cp=asymptomatic}$& -4& -2& -1\\
            $\bm{restecg=normal}$& 2& 1& 0\\
            $\bm{thal=reversable defect}$& 2& 4& 5\\
            $\bm{ca=0.0}$& 2& 0& -1\\
            $\bm{0.0 \leq oldpeak < 2.2}$& 2& 1& 0\\
            $+$ \space $\bm{bias}$& -14& -13& -13\\
            \hline
            Score:& = ....& = ....& = ....\\\hline
        \end{tabular}
        \label{tab:example_mcriskslim}
        }
    \end{subtable}
\end{table}

\textbf{Utility.} One of the advantages of our method is the ease of use. MISS generates a multiclass scoring system that can be used to conduct multiclass classification on paper without the need to use a computer. The simplicity, interpretability and good calibration of the model give perspective on integrating such scoring systems in many sensitive domains like healthcare or criminal justice. However, there are many scoring systems with decent \textit{optimality gap} ($\leq$ 0.2) that can be created with our method. Every model resulting from the 5-Fold CV is going to be different. The problem of choosing the best one for real-life implementation should be conducted with the assistance of domain experts.  

Our scoring system can be easily transformed into well-calibrated class probabilities with a \textit{softmax} function that can translate scores into probabilities. In order to avoid the need to compute \textit{softmax}, a table of all possible scores with associated probabilities can be generated. In this way, the usage of MISS takes only a summation of small integers. The drawback of such an approach is that the created table would contain many entries as the number of possible combinations grows rapidly with the number of coefficients. MISS can also be utilized as a method for generating binary classification scoring systems, which is an interesting alternative to existing approaches like RiskSLIM. Our method finds points for every class, thus, the contribution of every feature to the final per-class scores can be analyzed.

There are disadvantages of our method. Firstly, the performance of MISS depends on the binarization procedure applied to the dataset. Integrating feature binarization within the optimization algorithm seems a promising approach, however, it increases the complexity of the optimized problem. 
Finding the optimal number of bins and optimal threshold values should be the subject of additional research. Finally, the computational complexity of our method is high. With every additional class, the optimized problem gets harder as it requires finding $R^{max}$ coefficients for the new class.

\section{Conclusion}
In this paper, we presented an IP-based method to create multiclass scoring systems. MISS achieves performance on par with other interpretable machine learning algorithms and produces a single, sparse, well-calibrated, easy-to-use model, which allows classifying based on values of binary features and the sum of small integer coefficients. The utility of our method is broadened by the fact that it works for binary tasks as well. We presented methods for improving the performance of the model and RFA, a dimensionality reduction technique lowering the \textit{optimality gap}. In the future, we would like to extend MISS to multi-label classification problems and include binarization in the model optimization process as the model performance is highly dependent on the discretization techniques.

\section*{Acknowledgements}
This research was funded in whole or in part by National Science Centre, Poland 2023/49/N/ST6/01841. For the purpose of Open Access, the author has applied a CC-BY public copyright licence to any Author Accepted Manuscript (AAM) version arising from this submission. This work is supported by the European Union’s Horizon 2020 research and innovation programme under grant agreement Sano No 857533 and the International Research Agendas programme of the Foundation for Polish Science, co-financed by the European Union under the European Regional Development Fund.


\begin{thebibliography}{99}
\bibitem{akiba2019optuna}Akiba, T., et al. A next-generation hyperparameter optimization framework. {\em Proceedings Of The 25th ACM SIGKDD International Conference On Knowledge Discovery \& Data Mining}. pp. 2623-2631 (2019)
\bibitem{billiet2017interval}Billiet, L., Van Huffel, S. \& Van Belle, V. Interval coded scoring extensions for larger problems. {\em 2017 IEEE Symposium On Computers And Communications (ISCC)}. pp. 198-203 (2017)
\bibitem{antman2000timi}Antman, E., et al. The TIMI risk score for unstable angina/non–ST elevation MI: a method for prognostication and therapeutic decision making. {\em Jama}. \textbf{284}, 835-842 (2000)
\bibitem{segmentation_2021}AV Customer segmentation - Analytics Vidhya. {\em Kaggle}.
\bibitem{burgess1928factors}Burgess, E. Factors determining success or failure on parole. {\em The Workings Of The Indeterminate Sentence Law And The Parole System In Illinois}. pp. 221-234 (1928)
\bibitem{chen2016xgboost}Chen, T. \& Guestrin, C. Xgboost: A scalable tree boosting system. {\em Proceedings Of The 22nd ACM SIGKDD International Conference On Knowledge Discovery And Data Mining}. pp. 785-794 (2016)
\bibitem{cplex2009v12}Cplex, I. V12. 1: User’s Manual for CPLEX. {\em International Business Machines Corporation}. \textbf{46}, 157 (2009)
\bibitem{detrano1989international}Detrano, R., et al. International application of a new probability algorithm for the diagnosis of coronary artery disease. {\em The American Journal Of Cardiology}. \textbf{64}, 304-310 (1989)
\bibitem{duwe2016sacrificing}Duwe, G. \& Kim, K. Sacrificing accuracy for transparency in recidivism risk assessment: The impact of classification method on predictive performance. {\em Corrections}. \textbf{1}, 155-176 (2016)
\bibitem{finlay2012credit}Finlay, S. Credit scoring, response modeling, and insurance rating: a practical guide to forecasting consumer behavior. (Springer, 2012)
\bibitem{fisher2011uci}Fisher, R. UCI machine learning repository: Iris data set.  (2011)
\bibitem{forina1991uci}Forina, M. Uci machine learning repository wine dataset. {\em Institute Of Pharmaceutical And Food Analysis And Technologies}. (1991)
\bibitem{gage2001validation}Gage, B., et al. Validation of clinical classification schemes for predicting stroke: results from the National Registry of Atrial Fibrillation. {\em Jama}. \textbf{285}, 2864-2870 (2001)
\bibitem{guo2017calibration}Guo, C., Pleiss, G., Sun, Y. \& Weinberger, K. On calibration of modern neural networks. {\em International Conference On Machine Learning}. pp. 1321-1330 (2017)
\bibitem{guyon2002gene}Guyon, I., Weston, J., Barnhill, S. \& Vapnik, V. Gene selection for cancer classification using support vector machines. {\em Machine Learning}. \textbf{46}, 389-422 (2002)
\bibitem{Hurdman2012}Hurdman, J., et al. ASPIRE registry: Assessing the Spectrum of Pulmonary hypertension Identified at a REferral centre. {\em European Respiratory Journal}. \textbf{39}, 945-955 (2012,4)
\bibitem{kallus2018residual}Kallus, N. \& Zhou, A. Residual unfairness in fair machine learning from prejudiced data. {\em International Conference On Machine Learning}. pp. 2439-2448 (2018)
\bibitem{kelley1960cutting}Kelley, J. The cutting-plane method for solving convex programs. {\em Journal Of The Society For Industrial And Applied Mathematics}. \textbf{8}, 703-712 (1960)
\bibitem{kumar2019verified}Kumar, A., et al. Verified Uncertainty Calibration. {\em Advances In Neural Information Processing Systems}. \textbf{32} (2019)
\bibitem{liu2022fasterrisk}Liu, J., et al. FasterRisk: Fast and Accurate Interpretable Risk Scores. {\em Proceedings Of Neural Information Processing Systems}. (2022)
\bibitem{naeini2015binary}Naeini, M., Cooper, G. \& Hauskrecht, M. Binary classifier calibration using a Bayesian non-parametric approach. {\em Proceedings Of The 2015 SIAM International Conference On Data Mining}. pp. 208-216 (2015)
\bibitem{pajor2022effect}Pajor, A., et al. Effect of Feature Discretization on Classification Performance of Explainable Scoring-Based Machine Learning Model. {\em International Conference On Computational Science}. pp. 92-105 (2022)
\bibitem{scikit-learn}Pedregosa, F., et al. Scikit-learn: Machine Learning in Python. {\em Journal Of Machine Learning Research}. \textbf{12} pp. 2825-2830 (2011)
\bibitem{proencca2020interpretable}Proença, H. \& Leeuwen, M. Interpretable multiclass classification by MDL-based rule lists. {\em Information Sciences}. \textbf{512} pp. 1372-1393 (2020)
\bibitem{rouzot2022learning}Rouzot, J., Ferry, J. \& Huguet, M. Learning Optimal Fair Scoring Systems for Multi-Class Classification. {\em ICTAI 2022-The 34th IEEE International Conference On Tools With Artificial Intelligence}. (2022)
\bibitem{smith1988using}Smith, J., et al. Using the ADAP learning algorithm to forecast the onset of diabetes mellitus. {\em Proceedings Of The Annual Symposium On Computer Application In Medical Care}. pp. 261 (1988)
\bibitem{street1993nuclear}Street, W., et al. Nuclear feature extraction for breast tumor diagnosis. {\em Biomedical Image Processing And Biomedical Visualization}. \textbf{1905} pp. 861-870 (1993)
\bibitem{sultana2018predicting}Sultana, J. \& Jilani, A. Predicting breast cancer using logistic regression and multi-class classifiers. {\em International Journal Of Engineering \& Technology}. \textbf{7}, 22-26 (2018)
\bibitem{than2014development}Than, M., et al. Development and validation of the E mergency D epartment A ssessment of C hest pain S core and 2 h accelerated diagnostic protocol. {\em Emergency Medicine Australasia}. \textbf{26}, 34-44 (2014)
\bibitem{ustun2016supersparse}Ustun, B. \& Rudin, C. Supersparse linear integer models for optimized medical scoring systems. {\em Machine Learning}. \textbf{102}, 349-391 (2016)
\bibitem{ustun2019learning}Ustun, B. \& Rudin, C. Learning Optimized Risk Scores. {\em J. Mach. Learn. Res.}. \textbf{20}, 1-75 (2019)
\bibitem{verwer2019learning}Verwer, S. \& Zhang, Y. Learning optimal classification trees using a binary linear program formulation. {\em Proceedings Of The AAAI Conference On Artificial Intelligence}. \textbf{33}, 1625-1632 (2019)
\bibitem{wang2022pursuit}Wang, C., Han, B., Patel, B. \& Rudin, C. In pursuit of interpretable, fair and accurate machine learning for criminal recidivism prediction. {\em Journal Of Quantitative Criminology}. pp. 1-63 (2022)
\bibitem{xu2020multi}Xu, J., et al. A multi-class scoring system based on CT features for preoperative prediction in gastric gastrointestinal stromal tumors. {\em American Journal Of Cancer Research}. \textbf{10}, 3867 (2020)
\bibitem{zadrozny2002transforming}Zadrozny, B. \& Elkan, C. Transforming classifier scores into accurate multiclass probability estimates. {\em Proceedings Of The Eighth ACM SIGKDD International Conference On Knowledge Discovery And Data Mining}. pp. 694-699 (2002)
\bibitem{zeng2017interpretable}Zeng, J., et al. Interpretable classification models for recidivism prediction. {\em Journal Of The Royal Statistical Society: Series A (Statistics In Society)}. 
\bibitem{haoran2021checklists}Zhang, H., et al. Learning Optimal Predictive Checklists. {\em Advances In Neural Information Processing Systems}. \textbf{34} pp. 1215-1229 (2021)\textbf{180}, 689-722 (2017)
\end{thebibliography}

\pagebreak
\onecolumn
\appendix
\section{On scoring systems}
\label{appendix:scoring_systems}
\begin{figure}[h!]
\centering
  \includegraphics[width=10cm]{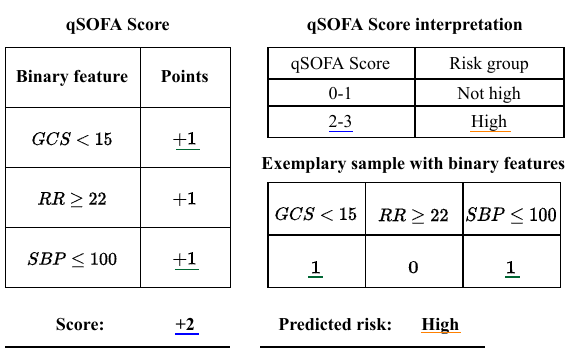}

\caption{Example of qSOFA scoring system measuring the risk of in-hospital mortality with suspected sepsis infection. The binary features used by this scoring system are: 1. Altered mental status Glasgow Coma Scale (GCS), 2. Respiratory rate (RR), 3. Systolic Blood Pressure (SBP). } 
\label{fig:qsofa_score}
\end{figure}

There are many examples of binary scoring systems utilized in healthcare. For instance qSOFA\footnote{Seymour, C., Liu, V., Iwashyna, T., Brunkhorst, F., Rea, T., Scherag, A., Rubenfeld, G., Kahn, J., Shankar-Hari, M., Singer, M. \& Others Assessment of clinical criteria for sepsis: for the Third International Consensus Definitions for Sepsis and Septic Shock (Sepsis-3). {\em Jama}. \textbf{315}, 762-774 (2016)} (Quick SOFA, Figure \ref{fig:qsofa_score}) is a method for assessing the risk of in-hospital mortality for patients with suspected sepsis infection. This scoring system uses three binary features. If at least two features are positive (score $\geq$ 2), the predicted risk is \textit{High}. Our method, MISS, can generate scoring systems for binary and multiclass tasks. In the MISS model, the predicted class is the one with the maximum score.

\section{Model customization} 
\label{appendix:customization}

One of the benefits of using IP solver to find a solution to (\ref{eq:risk_score_problem}) is that the user can specify MISS requirements with the addition of custom constraints. We present examples of such customization in Table \ref{tab:custom_constraints}. These methods allow the creation of models that obey domain-specific requirements and can improve the fairness and validity of the model \cite{haoran2021checklists}.

\begin{table}[h!]
    \caption{Examples of custom constraints that can be embedded into solved IP problem of MISS.}
    \label{tab:custom_constraints}
    \centering
    \begin{tabular}{ll}
    \hline
         \textbf{Requirement} & \textbf{Example}  \\ \hline
         Model size & Use maximally 5 features \\ 
         Prediction & If \textit{hypertension} predict \textit{critical heart failure} \\ 
         Structure & Include \textit{is\_male} if \textit{diabetes} is in the model \\ 
         Scoring & Coefficient of \textit{BMI} is the highest for class \textit{cancer} \\
         Sparsity & Use 10 to 15 coefficients \\ 
         Group sparsity & Include $g \leq age \leq h$ at most 2 times\\ \hline
    \end{tabular}
\end{table}

\section{Datasets}
\label{appendix:datasets}
In what follows, we give more details on the datasets used in this paper. We use three binary and five multiclass datasets. All of the datasets (apart from \dsFont{ph\_binary} and \dsFont{ph}) are publicly available. All numerical features were subject to binarization. We one-hot encode all categorical features.

\dsFont{diabetes} is a binary dataset from the National Institute of Diabetes and Digestive and Kidney Diseases containing information about 768 females at least 21 years old and of Pima Indian heritage \cite{smith1988using}. All available features are numerical. The classification task of this dataset is to predict whether a woman has tested positive for diabetes.

\dsFont{breast\_cancer} is a binary Breast Cancer Wisconsin (Diagnostic) dataset containing 30 real-valued, positive features \cite{street1993nuclear}. The features of 569 samples are describing the breast cell nuclei characteristics extracted from the specialized image of a fine needle aspirate of a breast mass. This dataset allows the prediction of whether the patient has breast cancer.

\dsFont{ph\_binary} is a binary dataset containing 353 samples of patients who underwent Cardiac Magnetic Resonance Imaging (MRI) and Main Pulmonary Artery (MPA) MRI at Sheffield Teaching Hospitals. Additionally, patients underwent Right Heart Catheterization (RHC) within 24-hours to measure the pressure inside the MPA. The samples are part of the ASPIRE Registry (Assessing the Severity of Pulmonary Hypertension In a Pulmonary Hypertension REferral Centre) \cite{Hurdman2012}. There are 35 numerical features which are the measurements from MRI images and videos - e.g. left ventricle ejection fraction, MPA systolic area, right ventricle stroke volume. The goal is to predict whether the patient suffers from Pulmonary Hypertension (PH). PH, in the case of this dataset, was positive, if the pressure measured during RHC was equal or above 25 mmHg.  

\dsFont{iris} is a popular 3-class problem dataset comprising of 150 samples with 4 numerical features of 3 iris species \cite{fisher2011uci}. The samples are categorized into one of the three species: setosa, versicolor and virginica.

\dsFont{wine} is another popular dataset with 3 classes \cite{forina1991uci}. The samples contain 13 numerical, real-valued features collected from a chemical analysis (e.g. color intensity, amount of alcohol) of wines grown from 3 different cultivars.

\dsFont{heart} is a Heart Disease (HD) dataset with 8 categorical (e.g. sex, chest pain type) and 5 numerical (e.g. maximum heart rate) \cite{detrano1989international}. We discard 2 features - \textit{id} and the origin of study. Originally, the dataset comprises of 5 classes - \textit{No HD} and four stages of HD. Since the number of the latter stages is small, we combine samples with stages 2 \& 3 and 4 \& 5 to create \textit{Mediocre HD} and \textit{Severe HD} classes respectively.

\dsFont{ph} contains the same samples as \dsFont{ph\_binary} \cite{Hurdman2012}. However, in this case, the predicted variable is not binary. Instead of distinguishing between PH and no PH, the problem is multiclass. The PH class was divided into two separate classes: Pulmonary Arterial Hypertension (PAH) - the most common type of PH in the dataset (142 samples) and all other types of PH (e.g. PH resulting from Chronic Obstructive Pulmonary Disease). This resulted in the creation of 3 class problem. All features are numerical and are subject to feature binarization.

\dsFont{segmentation} is the 4-class dataset from the Analytics Vidhya Janatahack competition \cite{segmentation_2021}. The goal is to assign clients into one of the four segments - \textit{A, B, C, D} based on their profession, age, gender etc. The dataset contains 3 numerical and 6 categorical features. We use only the training set from the challenge, as the publicly available testing set does not contain labels.

\section{Hyperparameter optimization}
\label{appendix:hyperparameter}
In this section, we provide information on the search grids used during hyperparameter optimization for every tested method. All parameters that are not mentioned are set to default values.

\methodFont{RF} - we use the implementation from the sckit-learn library \cite{scikit-learn} and set the following search grid: $criterion \in \{"gini", "entropy", "log\_loss"\}$, $n\_estimators \in \{5, ..., 100\}$, $min\_samples\_split \in \{2, ..., 10\}$, $min\_samples\_leaf \in \{2, ..., 5\}$.

\methodFont{XGB} - we use the XGBoost library \cite{chen2016xgboost} and set the search grid to: $max\_depth \in \{1, ..., 7\}$, $n\_estimators \in \{5, ..., 100\}$.
 
\methodFont{UNIT} - we train the L1-penalized Logistic Regression from the scikit-learn library with \textit{saga} optimizer and maximum number of iterations set to 1000. After the model is trained we set the intercept value to 0 and assign +1, 0 or -1 to coefficient values depending on their sign. We sample the parameter $C \in [10^{-4}, 10]$ from the log domain.
 
\methodFont{LR} - we train the L1-penalized Logistic Regression from the scikit-learn library with \textit{saga} optimizer and maximum number of iterations set to 1000. We sample the parameter $C \in [10^{-4}, 10]$ from the log domain.
 
\methodFont{DT} - we use Decision Trees implementation from the scikit-learn library and set the following search grid: $criterion \in \{"gini", "entropy", "log\_loss"\}$, $min\_samples\_split \in \{2, ..., 10\}$, $min\_samples\_leaf \in \{2, ..., 5\}$.
 
\methodFont{RuleList} - we use the code provided in \cite{proencca2020interpretable}, set \textit{task} to \textit{prediction} and \textit{target\_model} to \textit{categorical}. We vary $max\_depth \in \{2, ..., 7\}$ and $beam\_width \in \{50, ..., 150\}$.

\methodFont{FasterRisk} - we use the code provided in \cite{liu2022fasterrisk} train FasterRisk in the OvR manner, set \textit{sparsity} to 5 and all other parameters to default values.
 
\methodFont{OvR-RiskSLIM} - we use the code provided in \cite{ustun2019learning} to train binary RiskSLIM and to implement the One-vs-Rest RiskSLIM for multiclass classification. We set the parameters  of \methodFont{OvR-RiskSLIM} in the following way: max coefficient of 5, max intercept of 20, max model size of 10, penalization term of $10^{-6}$ and a timeout of 20 minutes per each of $K$ models
 
\methodFont{MISS} is trained with max coefficient of 5 ($\lambda_{j,k} \in \{-5, ..., 5\}$ for every $j \in \{1, ..., D\}$ and $k \in \{0, ..., K-1\}$ ),  setting max bias to 20 ($[\lambda_{00}, ..., \lambda_{0K-1}] \in \{-20, ..., 20\}^{K-1}$), maximum model size of $R^{max}=5$ ($B \leq 5$) and small sparsity penalization term $C_{0}=10^{-6}$ with a timeout of 90 minutes.

Additionally, we optimize hyperparameters of \textit{kbins} discretizers \cite{pajor2022effect}. For \textit{kbins} method we vary the $strategy \in \{"uniform", "quantile", "kmeans"\}$ and the $n\_bins \in \{2, 3, 4\}$. For \textit{MDLP} discretizers we set $precision = 2$. For both discretization methods, all numerical features are discretized in a non-overlapping manner and all other parameters are set to default values.

\section{Examples of multiclass scoring systems}
\label{appendix:examples}
In what follows, we present the examples of scoring systems generated by MISS for all datasets ($R^{max} \leq 5$ and $R^{max} \leq 10$).

\begin{table}[h!]
\caption{Examples of scoring systems generated by MISS with $R^{max} \leq 5$ (a) and $R^{max} \leq 10$ (b) for \dsFont{diabetes} dataset (AUC=0.86).}
\label{tab:example_diabetes}

\centering
    \begin{subtable}[t]{0.49\textwidth}
        \caption{}
        \centering
        \adjustbox{width =\textwidth}{
        \begin{tabular}{l*{2}{c}}\hline
        &  \multicolumn{2}{c}{Class} \\\cline{2-3}
                Binary feature &  \textbf{No Diabetes}& \textbf{Diabetes}\\\hline
        $\bm{Pregnancies < 6.0}$& 5& 4\\
        $\bm{Glucose < 100.0}$& 1& 0\\
        $\bm{144.0 \leq Glucose \leq 199.0}$& -2& 0\\
        $\bm{26.4 \leq BMI \leq 67.1}$& 3& 5\\
        $\bm{28.0 \leq Age \leq 81.0}$& 4& 5\\
        $+$\space $\bm{bias}$& -17& -19\\
        \hline
        Score:& = ....& = ....\\\hline
        \end{tabular}
        \label{tab:example_diabetesb}
        }
    \end{subtable}
    \begin{subtable}[t]{0.49\textwidth}
        \caption{}
        \centering
        \adjustbox{width =\textwidth}{
        \begin{tabular}{l*{2}{c}}\hline
        &  \multicolumn{2}{c}{Class} \\\cline{2-3}
                Binary feature &  \textbf{No Diabetes}& \textbf{Diabetes}\\\hline
        $\bm{Pregnancies < 6.0}$& 0& 5\\
        $\bm{6.0 \leq Pregnancies \leq 17.0}$& -5& 1\\
        $\bm{Glucose < 100.0}$& -4& -5\\
        $\bm{144.0 \leq Glucose \leq 199.0}$& -1& 1\\
        $\bm{23.0 \leq SkinThickness \leq 60.0}$& -1& -1\\
        $\bm{26.4 \leq BMI \leq 67.1}$& -5& -3\\
        $\bm{21.0 \leq Age < 28.0}$& 1& -1\\
        $\bm{28.0 \leq Age \leq 81.0}$& 1& 0\\
        $+$\space $\bm{bias}$& -14& -20\\
        \hline
        Score:& = ....& = ....\\\hline
        \end{tabular}
        \label{tab:example_diabetesa}
        }
    \end{subtable}
\end{table}

\begin{table}[h!]
\caption{Examples of scoring systems generated by MISS with $R^{max} \leq 5$ (a) and $R^{max} \leq 10$ (b) for \dsFont{breast\_cancer} dataset (AUC=0.99 and 1.0 respectively).}
\label{tab:example_breastcancer}
\centering
    \begin{subtable}[t]{0.49\textwidth}
        \caption{}
        \centering
        \adjustbox{width =\textwidth}{
        \begin{tabular}{l*{2}{c}}\hline
        &  \multicolumn{2}{c}{Class} \\\cline{2-3}
                Binary feature & \textbf{No Cancer}& \textbf{Cancer}\\\hline
        $\bm{696.25 \leq mean area \leq 2501.0}$& 0& -3\\
        $\bm{0.12 \leq mean concavity \leq 0.4268}$& 4& 2\\
        $\bm{0.0 \leq mean concave points < 0.05}$& -2& 0\\
        $\bm{185.2 \leq worst area < 693.4}$& -2& 0\\
        $\bm{0.0 \leq worst concavity < 0.21}$& 0& 2\\
        $+$\space $\bm{bias}$& -19& -20\\
        \hline
        Score:& = ....& = ....\\\hline
        \end{tabular}
        \label{tab:example_brestcancera}
        }
    \end{subtable}
    \begin{subtable}[t]{0.49\textwidth}
        \caption{}
        \centering
        \adjustbox{width =\textwidth}{
        \begin{tabular}{l*{2}{c}}\hline
        &  \multicolumn{2}{c}{Class} \\\cline{2-3}
                Binary feature &  \textbf{No Cancer}& \textbf{Cancer}\\\hline
        $\bm{13.37 \leq mean radius \leq 27.42}$& 4& 5\\
        $\bm{674.23 \leq mean area \leq 2501.0}$& 0& -1\\
        $\bm{0.0 \leq mean concavity < 0.06}$& 2& 1\\
        $\bm{0.0 \leq mean concave points < 0.05}$& 2& 5\\
        $\bm{7.93 \leq worst radius < 18.56}$& -4& -2\\
        $\bm{50.41 \leq worst perimeter < 117.34}$& -5& 5\\
        $\bm{117.34 \leq worst perimeter < 184.27}$& -5& 4\\
        $\bm{688.6 \leq worst area \leq 4254.0}$& 4& 1\\
        $\bm{0.0 \leq worst concavity < 0.23}$& 3& 4\\
        $\bm{0.0 \leq worst concave points < 0.15}$& -1& 1\\
        $+$\space $\bm{bias}$& -6& -19\\
        \hline
        Score:& = ....& = ....\\\hline
        \end{tabular}
        \label{tab:example_brestcancerb}
        }
    \end{subtable}
\end{table}

\begin{table}[h!]
\caption{Examples of scoring systems generated by MISS with $R^{max} \leq 5$ (a) and $R^{max} \leq 10$ (b) for \dsFont{ph\_binary} dataset (AUC=0.99 and 1.0 respectively).}
\label{tab:example_phbinary}
\centering
    \begin{subtable}[t]{0.49\textwidth}
        \caption{}
        \centering
        \adjustbox{width =\textwidth}{
        \begin{tabular}{l*{2}{c}}\hline
        &  \multicolumn{2}{c}{Class} \\\cline{2-3}
                Binary feature &  \textbf{No PH}& \textbf{PH}\\\hline
        $\bm{6.64 \leq diastolic\_fiesta \leq 25.64}$& -1& 0\\
        $\bm{21.0 \leq rvesv < 102.2}$& 5& -5\\
        $\bm{42.69 \leq rvef \leq 81.06}$& 0& -1\\
        $\bm{31.0 \leq rv\_dia\_mass \leq 223.0}$& -2& 0\\
        $\bm{112.0 \leq septal\_angle\_syst < 164.0}$& 3& -1\\
        $+$\space $\bm{bias}$& 6& 19\\
        \hline
        Score:& = ....& = ....\\\hline
        \end{tabular}
        \label{tab:example_phbina}
        }
    \end{subtable}
    \begin{subtable}[t]{0.49\textwidth}
        \caption{}
        \centering
        \adjustbox{width =\textwidth}{
        \begin{tabular}{l*{2}{c}}\hline
        &  \multicolumn{2}{c}{Class} \\\cline{2-3}
                Binary feature &  \textbf{No PH}& \textbf{PH}\\\hline
        $\bm{bias}$
        $\bm{8.23 \leq diastolic\_fiesta \leq 15.79}$& -1& 0\\
        $\bm{1.58 \leq rvesv\_index < 63.34}$& 5& -5\\
        $\bm{63.34 \leq rvesv\_index \leq 125.11}$& -5& 5\\
        $\bm{26.73 \leq rvef < 39.73}$& -5& -2\\
        $\bm{10.0 \leq rv\_dia\_mass < 33.66}$& 5& -5\\
        $\bm{33.66 \leq rv\_dia\_mass < 60.63}$& 3& -5\\
        $\bm{0.0 \leq rv\_mass\_index < 22.88}$& 0& 1\\
        $\bm{112.0 \leq septal\_angle\_syst < 139.75}$& 1& 0\\
        $\bm{167.5 \leq septal\_angle\_syst < 195.25}$& -5& 5\\
        $\bm{14.8 \leq pa\_qflowneg\_index \leq 60.36}$& 0& 1\\
        $+$\space $\bm{bias}$& -20& -2\\
        \hline
        Score:& = ....& = ....\\\hline
        \end{tabular}
        \label{tab:example_phbinb}
        }
    \end{subtable}
\end{table}

\begin{table}[h!]
\caption{Examples of scoring systems generated by MISS with $R^{max} \leq 5$ (a) and $R^{max} \leq 10$ (b) for \dsFont{iris} dataset (AUC=1.0).}
\label{tab:example_iris}
\centering
    \begin{subtable}[t]{0.49\textwidth}
        \caption{}
        \centering
        \adjustbox{width =\textwidth}{
        \begin{tabular}{l*{3}{c}}\hline
        &  \multicolumn{3}{c}{Class} \\\cline{2-4}
                Binary feature &  \textbf{setosa}& \textbf{versicolor}& \textbf{virignica}\\\hline
        $\bm{4.3 \leq sepal\_length < 6.1}$& 5& 0& -2\\
        $\bm{2.0 \leq sepal\_width < 2.9}$& -5& -3& 0\\
        $\bm{1.0 \leq petal\_length < 2.97}$& 5& -5& -5\\
        $\bm{2.97 \leq petal\_length < 4.93}$& -5& 3& -1\\
        $\bm{0.8 \leq petal\_width < 1.7}$& -5& 5& -1\\
        $+$ \space $\bm{bias}$& 2& 7& 12\\
        \hline
        Score:& = ....& = ....& = ....\\\hline
        \end{tabular}
        \label{tab:example_irisa}
        }
    \end{subtable}
    \begin{subtable}[t]{0.49\textwidth}
        \caption{}
        \centering
        \adjustbox{width =\textwidth}{
        \begin{tabular}{l*{3}{c}}\hline
        &  \multicolumn{3}{c}{Class} \\\cline{2-4}
                Binary feature &  \textbf{setosa}& \textbf{versicolor}& \textbf{virignica}\\\hline
        $\bm{4.3 \leq sepal\_length < 5.56}$& 0& -5& -3\\
        $\bm{5.56 \leq sepal\_length < 6.62}$& -1& 0& 5\\
        $\bm{6.62 \leq sepal\_length \leq 7.9}$& -5& -1& 5\\
        $\bm{2.0 \leq sepal\_width < 2.8}$& -5& 2& 5\\
        $\bm{2.8 \leq sepal\_width < 3.0}$& -5& 0& 2\\
        $\bm{3.4 \leq sepal\_width \leq 4.4}$& 0& -3& -2\\
        $\bm{1.0 \leq petal\_length < 3.21}$& 5& -5& -5\\
        $\bm{3.21 \leq petal\_length \leq 6.9}$& -5& -2& 4\\
        $\bm{0.1 \leq petal\_width < 0.8}$& 5& -5& -5\\
        $\bm{0.8 \leq petal\_width < 1.7}$& -5& 4& -2\\
        $+$ \space $\bm{bias}$& -16& -6& -15\\
        \hline
        Score:& = ....& = ....& = ....\\\hline
        \end{tabular}
        \label{tab:example_irisb}
        }
    \end{subtable}
\end{table}

\begin{table}[h!]
\caption{Examples of scoring systems generated by MISS with $R^{max} \leq 5$ (a) and $R^{max} \leq 10$ (b) for \dsFont{wine} dataset (AUC=1.0).}
\label{tab:example_wine}
\centering
    \begin{subtable}[t]{0.49\textwidth}
        \caption{}
        \centering
        \adjustbox{width =\textwidth}{
        \begin{tabular}{l*{3}{c}}\hline
        &  \multicolumn{3}{c}{Class} \\\cline{2-4}
                Binary feature &  \textbf{class 0}& \textbf{class 1}& \textbf{class 2}\\\hline
        $\bm{11.41 \leq alcohol < 12.76}$& -5& 5& 3\\
        $\bm{0.34 \leq flavanoids < 0.95}$& -5& -5& 5\\
        $\bm{2.31 \leq flavanoids \leq 3.93}$& 5& 0& -5\\
        $\bm{1.74 \leq color\_intensity < 3.46}$& -5& 5& -5\\
        $\bm{0.48 \leq hue < 0.79}$& -5& -5& 5\\
        $+$ \space $\bm{bias}$& -14& -13& -17\\
        \hline
        Score:& = ....& = ....& = ....\\\hline
        \end{tabular}
        \label{tab:example_winea}
        }
    \end{subtable}
    \begin{subtable}[t]{0.49\textwidth}
        \caption{}
        \centering
        \adjustbox{width =\textwidth}{
        \begin{tabular}{l*{3}{c}}\hline
        &  \multicolumn{3}{c}{Class} \\\cline{2-4}
                Binary feature &  \textbf{class 0}& \textbf{class 1}& \textbf{class 2}\\\hline
        $\bm{12.78 \leq alcohol \leq 14.83}$& 5& -5& -5\\
        $\bm{0.34 \leq flavanoids < 0.98}$& -5& -5& 5\\
        $\bm{1.58 \leq flavanoids < 2.31}$& 4& 4& -5\\
        $\bm{2.31 \leq flavanoids \leq 5.08}$& 5& -3& -5\\
        $\bm{1.28 \leq color\_intensity < 3.46}$& -5& 5& -5\\
        $\bm{0.96 \leq hue < 1.29}$& -3& 1& -5\\
        $\bm{1.27 \leq od280/od315\_of\_diluted\_wines < 2.06}$& -5& -4& 1\\
        $\bm{468.0 \leq proline < 755.0}$& -1& 2& 5\\
        $\bm{987.5 \leq proline \leq 1680.0}$& 5& -5& -5\\
        $+$ \space $\bm{bias}$& -20& -9& -12\\
        \hline
        Score:& = ....& = ....& = ....\\\hline
        \end{tabular}
        \label{tab:example_wineb}
        }
    \end{subtable}
\end{table}

\begin{table}[h!]
\caption{Examples of scoring systems generated by MISS with $R^{max} \leq 5$ (a) and $R^{max} \leq 10$ (b) for \dsFont{heart} dataset (AUC=0.88).}
\label{tab:example_heart}
\centering
    \begin{subtable}[t]{0.49\textwidth}
        \caption{}
        \centering
        \adjustbox{width =\textwidth}{
        \begin{tabular}{l*{3}{c}}\hline
        &  \multicolumn{3}{c}{Class} \\\cline{2-4}
                Binary feature &  \textbf{No HD}& \textbf{Mediocre HD}& \textbf{Severe HD}\\\hline
        $\bm{cp=asymptomatic}$& -1& 1& 2\\
        $\bm{slope=upsloping}$& 0& -1& -2\\
        $\bm{thal=reversable defect}$& -2& 0& 1\\
        $\bm{ca=0.0}$& -2& -4& -5\\
        $\bm{0.0 \leq oldpeak < 2.0}$& 5& 4& 3\\
        $+$ \space $\bm{bias}$& -10& -9& -9\\
        \hline
        Score:& = ....& = ....& = ....\\\hline
        \end{tabular}
        \label{tab:example_hearta}
        }
    \end{subtable}
    \begin{subtable}[t]{0.49\textwidth}
        \caption{}
        \centering
        \adjustbox{width =\textwidth}{
        \begin{tabular}{l*{3}{c}}\hline
        &  \multicolumn{3}{c}{Class} \\\cline{2-4}
                Binary feature &  \textbf{No HD}& \textbf{Mediocre HD}& \textbf{Severe HD}\\\hline
        $\bm{cp=asymptomatic}$& 0& 1& 1\\
        $\bm{cp=non-anginal}$& -3& -4& -4\\
        $\bm{restecg=normal}$& -3& -4& -5\\
        $\bm{exang=False}$& 5& 4& 4\\
        $\bm{slope=flat}$& 5& 5& 2\\
        $\bm{slope=upsloping}$& 0& -1& -5\\
        $\bm{thal=reversable defect}$& -5& -3& -2\\
        $\bm{ca=0.0}$& 0& -2& -4\\
        $\bm{ca=2.0}$& -1& -1& 0\\
        $\bm{113.95 \leq thalch < 139.92}$& 3& 3& 5\\
        $+$ \space $\bm{bias}$& 16& 18& 20\\
        \hline
        Score:& = ....& = ....& = ....\\\hline
        \end{tabular}
        \label{tab:example_heartb}
        }
    \end{subtable}
\end{table}

\begin{table}[h!]
\caption{Examples of scoring systems generated by MISS with $R^{max} \leq 5$ (a) and $R^{max} \leq 10$ (b) for \dsFont{ph} dataset (discrimination between no Pulmonary Hypertension (PH), Pulmonary Arterial Hypertension (PAH) and other types of PH) - AUC=0.73.}
\label{tab:example_ph}
\centering
    \begin{subtable}[t]{0.49\textwidth}
        \caption{}
        \centering
        \adjustbox{width =\textwidth}{
        \begin{tabular}{l*{3}{c}}\hline
        &  \multicolumn{3}{c}{Class} \\\cline{2-4}
                Binary feature &  \textbf{No PH}& \textbf{PAH}& \textbf{PH Other}\\\hline
        $\bm{2.74 \leq diastolic\_fiesta < 6.48}$& 2& 1& 0\\
        $\bm{108.83 \leq rvesv \leq 265.9}$& -5& 1& 0\\
        $\bm{49.65 \leq rvesv\_index \leq 128.6}$& 0& -1& 1\\
        $\bm{0.0 \leq rv\_mass\_index < 12.09}$& -3& -5& -3\\
        $\bm{118.0 \leq septal\_angle\_syst < 155.21}$& 5& 3& 4\\
        $+$ \space $\bm{bias}$& -10& -8& -9\\
        \hline
        Score:& = ....& = ....& = ....\\\hline
        \end{tabular}
        \label{tab:example_pha}
        }
    \end{subtable}
    \begin{subtable}[t]{0.49\textwidth}
        \caption{}
        \centering
        \adjustbox{width =\textwidth}{
        \begin{tabular}{l*{3}{c}}\hline
        &  \multicolumn{3}{c}{Class} \\\cline{2-4}
                Binary feature &  \textbf{No PH}& \textbf{PAH}& \textbf{PH Other}\\\hline
        $\bm{21.28 \leq rac\_fiesta \leq 76.72}$& 4& 3& 4\\
        $\bm{3.2 \leq diastolic\_fiesta < 7.75}$& 2& 2& 1\\
        $\bm{38.65 \leq rvedv\_index < 64.11}$& 5& 4& 5\\
        $\bm{21.0 \leq rvesv < 82.39}$& -5& -3& -4\\
        $\bm{1.58 \leq rvesv\_index < 48.43}$& -2& -3& -3\\
        $\bm{48.43 \leq rvesv\_index < 95.29}$& 5& 4& 5\\
        $\bm{10.0 \leq rv\_dia\_mass < 25.0}$& 2& 1& 2\\
        $\bm{112.0 \leq septal\_angle\_syst < 149.0}$& 5& 3& 4\\
        $\bm{180.0 \leq septal\_angle\_syst \leq 223.0}$& -5& 4& 3\\
        $\bm{146.0 \leq septal\_angle\_diast \leq 180.0}$& 0& 0& 1\\
        $+$ \space $\bm{bias}$&  -17& -16& -16\\
        \hline
        Score:& = ....& = ....& = ....\\\hline
        \end{tabular}
        \label{tab:example_phb}
        }
    \end{subtable}
\end{table}

\begin{table}[h!]
\caption{Examples of scoring systems generated by MISS with $R^{max} \leq 5$ (a) and $R^{max} \leq 10$ (b) for \dsFont{segmentation} dataset (AUC=0.75 and 0.74 respectively).}
\label{tab:example_segmentation}
\centering
    \begin{subtable}[t]{0.49\textwidth}
        \caption{}
        \centering
        \adjustbox{width =\textwidth}{

        \begin{tabular}{l*{4}{c}}\hline
        &  \multicolumn{4}{c}{Class} \\\cline{2-5}
                Binary feature &  \textbf{A}& \textbf{B}& \textbf{C}& \textbf{D}\\\hline
        $\bm{Profession=Artist}$& -3& -3& -2& -5\\
        $\bm{Profession=Healthcare}$& 3& 4& 5& 5\\
        $\bm{Spending\_Score=Low}$& 3& 2& 1& 4\\
        $\bm{18.0 \leq Age < 23.0}$& 1& 4& 3& 5\\
        $\bm{43.0 \leq Age < 73.5}$& -1& 0& 0& -1\\
        $+$ \space $\bm{bias}$& 18& 18& 18& 17\\
        \hline
        Score:& = ....& = ....& = ....& = ....\\\hline
        \end{tabular}
        \label{tab:example_segmentationa}
        }
    \end{subtable}
    \begin{subtable}[t]{0.49\textwidth}
        \caption{}
        \centering
        \adjustbox{width =\textwidth}{
        \begin{tabular}{l*{4}{c}}\hline
        &  \multicolumn{4}{c}{Class} \\\cline{2-5}
                Binary feature &  \textbf{A}& \textbf{B}& \textbf{C}& \textbf{D}\\\hline
        $\bm{Ever\_Married=No}$& -5& -5& -5& -5\\
        $\bm{Graduated=No}$& 4& 4& 4& 4\\
        $\bm{Graduated=Yes}$& -5& -5& -4& -5\\
        $\bm{Profession=Artist}$& 4& 5& 5& 3\\
        $\bm{Spending\_Score=Low}$& -4& -5& -5& -3\\
        $\bm{Var\_1=Cat\_4}$& 2& 1& 0& 2\\
        $\bm{18.0 \leq Age < 24.0}$& 1& 3& 3& 5\\
        $\bm{43.0 \leq Age < 74.0}$& -3& -2& -2& -3\\
        $\bm{3.0 \leq Family\_Size \leq 9.0}$& -1& -1& 0& -1\\
        $+$ \space $\bm{bias}$& -19& -19& -20& -20\\
        \hline
        Score:& = ....& = ....& = ....& = ....\\\hline
        \end{tabular}
        \label{tab:example_segmentationb}
        }
    \end{subtable}
\end{table}

\end{document}